\pdfoutput=1
\documentclass[11pt]{article}

\usepackage[final]{acl} 
\usepackage{times}
\usepackage{latexsym}
\usepackage{amsfonts}
\usepackage{amsmath}
\usepackage[T1]{fontenc}
\usepackage[utf8]{inputenc}
\usepackage{microtype}
\usepackage{inconsolata}
\usepackage{array,booktabs,siunitx}
\usepackage{float}
\usepackage{graphicx}
\usepackage{subcaption}
\usepackage{tablefootnote}
\usepackage{hyperref}
\usepackage{booktabs}
\usepackage{tcolorbox}
\newcolumntype{T}[1]{S[table-format=#1]}
\usepackage[skip=0.333\baselineskip]{caption}

\newcommand\scalemath[2]{\scalebox{#1}{\mbox{\ensuremath{\displaystyle #2}}}}

\title{Designing Informative Metrics for Few-Shot Example Selection}

\author{
  \hspace{3pt}Rishabh Adiga$^{\dagger}$\hspace{12pt} Lakshminarayanan Subramanian$^\ddagger$\hspace{12pt} Varun Chandrasekaran$^\dagger$  \vspace{0.3cm}\\
    \text{$^\dagger$University of Illinois Urbana-Champaign, IL}\\
    \text{$^\ddagger$New York University, NY}\\
    \text{\{radiga2, varunc\}@illinois.edu} \\
    \text{lakshmi@nyu.edu}
}

\begin{document}
\maketitle
\begin{abstract}

Pretrained language models (PLMs) have shown remarkable few-shot learning capabilities when provided with properly formatted examples. However, selecting the ``best'' examples remains an open challenge. We propose a complexity-based prompt selection approach for sequence tagging tasks. This approach avoids the training of a dedicated model for selection of examples, and instead uses certain metrics to align the syntactico-semantic complexity of test sentences and examples. We use both sentence- and word-level metrics to match the complexity of examples to the (test) sentence being considered. Our results demonstrate that our approach extracts greater performance from PLMs: it achieves state-of-the-art performance on few-shot NER, achieving a 5\% absolute improvement in F1 score on the CoNLL2003 dataset for \texttt{GPT-4}. We also see large gains of upto 28.85 points (F1/Acc.) in smaller models like \texttt{GPT-j-6B}.\footnote{Code and data used for the experiments in this paper is available at \url{https://github.com/RishabhAdiga/Complexity-based-prompt-retrieval.git}.}
\end{abstract}
\section{Introduction}
\label{sec:intro}

Pretrained language models (PLMs) have demonstrated impressive few-shot learning capabilities when prompted with examples for a particular task~\citep{brown2020language}. In few-shot prompting, $k$ examples that demonstrate the goal of the downstream task along with the test sample (collectively called the $k$-shot prompt) are provided to the PLM. However, the effectiveness of PLMs in few-shot learning depends on the ``quality'' of these prompts; achieving this quality is challenging due to the need for unambiguous instruction~\citep{sclar2024quantifying}, which is further complicated by the task of selecting appropriate examples.
Appropriately selected examples will ensure that the PLM can generalize effectively and perform well across diverse natural language processing (NLP) tasks.

Recent work has shown that specific examples provided in-context can significantly impact the performance of PLMs~\citep{min2022rethinking}.~\citet{Sorensen_2022} notes that specific example combinations lead to better generalization, and connects this with the informativeness of these examples. This suggests that surfacing ``specific'' properties of prompts result in improved performance. Most prior work for prompt selection has either trained a prompt retriever to select examples, or used PLMs themselves to label examples~\citep{rubin2022learning} to serve as a good prompt. However, both approaches require additional computation and task specific training.~\citet{liu-etal-2022-makes} use nearest neighbours algorithms to select examples. However, this approach solely relies on embedding space properties, and does not consider other metrics which model label distribution such as entropy, leading to sub-optimal performance in some tasks.

In this work, we propose a new approach for example selection for sequence tagging tasks. These tasks are important because they enable the extraction of structured information from unstructured text data; this serves as a fundamental building block for various NLP tasks.  
Our approach, {\em complexity-based prompt retrieval (CP retrieval)}, aims to select examples from the training data whose ``properties'' align with that of the test sentence. To this end, we use sentence- and word-level complexity measures to match prompt examples to evaluation sentences. 

We conduct our experiments on named entity recognition (NER), part-of-speech (POS) tagging, and sentence chunking. We also create a new dataset involving tagging of Contextual Integrity (CI) parameters for privacy policies. We show that CP retrieval improves accuracy significantly in all cases. 

The contributions of this work are as follows: (1) We propose CP retrieval to select prompts based on sentence and word level complexity metrics (\S~\ref{sec:methodology}); and (2) We demonstrate improved few-shot accuracy with CP retrieval. Our approach achieves state-of-the-art performance in few-shot NER with a 5\% absolute improvement in F1 score on the CoNLL2003 dataset for \texttt{GPT-4} and upto 28.85 points (F1/Accuracy) in smaller models (\S~\ref{sec:results}).

\section{Related Work}

To perform our experiments, we use the {\em Structured Prompting} paradigm introduced by~\citet{blevins2023prompting} as our baseline for named entity recognition (NER), part-of-speech (POS) tagging, and sentence chunking. These are sequence tagging tasks---those that assign labels or tags to tokens or other units of text. This involves iteratively feeding the PLM a word and its predicted label to get it to label the next word. PLMs are also provided $k$ pairs of (sentence, tagged sentence) as the task examples in a $k$-shot prompt. Experiments by~\citet{brown2020language} show that PLMs can effectively solve these tasks with just a few demonstrations using this technique. We consider another task involving annotating contextual integrity (CI) parameters in privacy policies~\citep{Shvartzshnaider2019GoingAT} ; we use a generic few-shot prompt as the baseline for this task as shown in Appendix~\ref{app:prompts} (Fig.~\ref{tab:prompt3}).

~\citet{liu-etal-2022-makes} have explored strategies for selecting examples to leverage \texttt{GPT-3}'s few-shot learning capabilities. They analyzed the sensitivity of \texttt{GPT-3}'s performance to randomly sampled examples and found significant variability (a phenomenon also explained by~\citet{min2022rethinking}). ~\citet{liu-etal-2022-makes} proposed a retrieval based approach called KATE that selects semantically similar examples by using $k$-nearest neighbors (in the embedding space of pre-trained sentence encoders like \texttt{BERT} and \texttt{RoBERTa}) which are used as the examples for the $k$-shot prompt. They found that KATE substantially improved \texttt{GPT-3}'s results over random sampling on natural language understanding and generation tasks. 

\section{Methodology}
\label{sec:methodology}

Our core idea is simple: {\em to select examples that closely resemble the test sample in both semantics and length while ensuring a diverse range of labels for the task.} We measure the ``complexity'' of a candidate example using 3 different sentence- and word-level metrics: (a) normalized sentence similarity score, (b) normalized smoothed length similarity, and (c) normalized label entropy.

\noindent\textbf{Problem Setup:} Assume we have a training dataset of the form $D =\{z_i = (x_i, y_i)\}_{i=1}^n$ and a
test example $x_{test}$, our goal is to use a complexity score to select a subset of examples $S$ of the training dataset ($S \subset D$) s.t. $|S|=k$. 

\subsection{Normalized Sentence Similarity}

For a given test sentence, we calculate its similarity scores with each candidate sentence for the $k$-shot prompt. We convert each of the sentences into sentence embeddings using the \texttt{all-MiniLM-L6-v2} sentence transformer (results do not vary drastically with alternate embedding methods). This produces an embedding size of 384 dimensions. To find the similarity score, we calculate cosine similarity between the embeddings.
$$\scalemath{0.83}{\texttt{Similarity}(x_i,x_{test}) = \texttt{cosine\_sim}(\texttt{emb}(x_i),\texttt{emb}(x_{test}))}$$
These values are normalized to get our final sentence similarity scores. Formally, this is denoted as \texttt{NormSentSimilarity}\footnote{Normalization here is N($s_i$) = $s_i$ / max($S$) where $S = \{s_1,\cdots, s_n\}$ and $s_i \in  \mathbb{R}$}.

\subsection{Normalized Smoothed Length Similarity}

This metric is based on the difference in sentence lengths between the candidate sentence and the test sentence. We use this metric since cosine similarity of sentence embeddings does not inherently factor sentence length. The smoothed length similarity (SLS) transforms the absolute length difference between two sentences using a sigmoid function to produce a smooth, tapered similarity curve rather than a hard threshold. The sigmoidal nature of SLS prevents stark drops in scores for small length divergences. This is important as, for few-shot prompting, some flexibility in length ranges is desirable to provide syntactic variety while maintaining coarse length matching. It is formulated as follows:
\[\scalemath{0.9}{\texttt{SLS}(x_i,x_{test})= (1+\exp({\dfrac{|\texttt{len}(x_i) - \texttt{len}(x_{test})|}{T}}))^{-1}}\]  
where $T$ controls the sigmoidal shape ($T$ = 3 in our experiments). Sentences with similar lengths achieve higher SLS scores, while increasingly divergent lengths taper down towards 0 in a continuous, non-binary fashion. These values are then normalized to provide our final Normalized Smoothed Length Similarity scores denoted as \texttt{NormSLS}\footnotemark[2].

\subsection{Normalized Label Entropy}

For a given task, let $\mathcal{Y}$ define the space of possible labels. For sentence $x_i$, let $y_i$ be the list of labels, and $\nu({y}^j_i)$ is the frequency of label $j \in \mathcal{Y}$ in list ${y}_i$. We can calculate $\Pr({y}^j_i) = \nu({y}^j_i)/\texttt{len}(x_i)$. Entropy is then calculated as: 
\[\text{H}({y}_i)=-\sum_{j \in \mathcal{Y}} \Pr({y}^j_i) \log_2 \Pr({y}^j_i)\]

The intuition is that sentences where labels are more skewed provide less information gain for in-context learning compared to flatter distributions. Entropy can be used to quantify that skew difference. We normalize these entropy values to get our final \texttt{NormEntropy}\footnotemark[2] scores.

\subsection{Complexity Score}
\label{sec:ComplexityScore}
We propose a complexity score to align the syntactico-semantic complexity of prompts and examples. The three component metrics are weighted and summed to produce the final complexity score:
\begin{align*}
\texttt{CS}(z_{i}, x_{test}) =  w_1*\texttt{NormSLS}(x_{i},x_{test})\\
 + w_2*\texttt{NormEntropy}(y_{i}) \\
 + w_3*\texttt{NormSentSimilarity}(x_{i}, x_{test})
\end{align*}

We use this score to select the $k$ highest scoring train sentences for a $k$-shot prompt. $w_1$, $w_2$, $w_3$ are weights that are set using grid search to optimize scores on the development set for each sequence tagging task. We found that $(w_1,w_2,w_3)$ = $(0.25,0.25,0.5)$ is the best set for NER, $(w_1,w_2,w_3)$ = $(0.2,0.1,0.7)$ is the best set for Chunking, $(w_1,w_2,w_3)$ = $(0.1,0.1,0.8)$ is the best set for POS Tagging and $(w_1,w_2,w_3)$ = $(0.1,0.1,0.8)$ is the best set for CI tagging. We see that the above set of values indicate that sentence similarity is most important. This is likely due to the presence of extremely similar examples in the test and training set.

\section{Experimental Setup}

\noindent{\bf Tasks:} We perform our experiments on 3 different sequence tagging tasks namely NER, POS tagging, and sentence chunking. Each task involves the annotation of tokens in a sequence with specific labels, facilitating the extraction of valuable information and enhancing language understanding. For all datasets, we use a random set of 1000 test samples.  For the CI task, we use 100 test samples.

\noindent{\em 1. NER:} This focuses on classifying named entities within a text. The entities can range from persons and organization to locations and dates. For our experiments, we use the widely recognized CoNLL2003 dataset~\citep{tjong-kim-sang-de-meulder-2003-introduction}.

\noindent{\em 2. POS Tagging:} This involves assigning grammatical categories (e.g., nouns, verbs, adjectives) to each word in a sentence.  We evaluate the POS tagging performance using the English Universal Dependencies (UD) treebank annotated on the GUM corpus~\citep{10.1007/s10579-016-9343-x}. It employs the UPOS tagset introduced by~\citet{nivre-etal-2020-universal}. 

\noindent{\em 3. Sentence Chunking:} This aims to partition sentences into non-overlapping syntactic units. For our experiments, we use the CoNLL2000 dataset~\citep{tjong-kim-sang-buchholz-2000-introduction} to frame chunking as a BIO tagging task.

\noindent{\em 4. CI parameters:} We additionally created a custom dataset that involves the sequence tagging task of CI parameters for privacy policies. For more details on this dataset, refer to Appendix~\ref{app:CI}.

\vspace{1mm}
\noindent{\bf Models:} We perform our experiments on the GPT-Neo series of models which has parameter counts from 125 million to 20 billion (\texttt{GPT-Neo-125M,GPT-Neo-1.3B, GPT-Neo-2.7B, GPT-j-6B, GPT-NeoX-20B}) and additionally 2 black box models (\texttt{Davinci-002} and \texttt{GPT-4}) for the sequence tagging tasks of NER, POS tagging and sentence chunking. For our custom dataset for CI parameters tagging, we use \texttt{Llama2} variants.
\begin{table*}[th!]
\centering
\resizebox{\textwidth}{!}{
\begin{tabular}{@{} l *{3}{T{1.2}T{1.2}} T{1.2}T{1.2} @{}}
\toprule
& \multicolumn{2}{c}{NER (F1)} 
& \multicolumn{2}{c@{}}{Chunking (F1)} 
& \multicolumn{2}{c@{}}{POS (Acc.)}\\
\cmidrule(lr){2-3} \cmidrule(l){4-5} \cmidrule(l){6-7}
& {Baseline} & {CP retrieval} & {Baseline} & {CP retrieval} & {Baseline} & {CP retrieval}\\
\midrule
\texttt{GPT-Neo-125M} & 12.46 & \textbf{29.21} & 24.28 & \textbf{40.33} & 55.53 & \textbf{71.09}\\
\texttt{GPT-Neo-1.3B}  & 31.49  & \textbf{52.44} & 25.44 & \textbf{44.19} & 65.30 & \textbf{75.58}\\
\texttt{GPT-Neo-2.7B}  & 25.73 & \textbf{41.77} & 28.50 & \textbf{53.32} & 64.68 & \textbf{76.19}\\
\texttt{GPT-j-6B}  & 25.88 &  \textbf{54.73} & 35.85 & \textbf{54.56} & 79.96 & \textbf{84.37}\\
\texttt{GPT-Neox-20B}  & 37.74 & \textbf{61.49} & 56.18 & \textbf{59.66} & 80.64 & \textbf{86.21}\\
\midrule
\texttt{Davinci-002} (Black Box) & 20.30 & \textbf{40.03} & 35.82 & \textbf{51.79} & 46.04\tablefootnote{Accuracy has changed with respect to the original paper for this model probably because of the the reasoning in \citet{chen2023chatgpts}} & \textbf{52.20}\\
\texttt{GPT-4} (Black Box) & 83.48* & \textbf{88.76} & 79.78 & \textbf{84.52} & 92.93 & \textbf{93.47}\\
\bottomrule
\end{tabular}}
\caption{ Results of using complexity scores for retrieval of the best $k$ examples (CP retrieval) for the few-shot prompt (here $k$ = 5). The baseline is in the structured prompting paradigm \citep{blevins2023prompting} with static prompt examples. {\bf CP retrieval significantly surpasses the baseline using linguistic structure alone for the prompt (details in Appendix~\ref{app:prompts}) in all cases.} Best values are in boldface. * in the table shows the previously achieved state-of-the-art performance on few-shot NER by~\citet{ashok2023promptner}.}
\label{tab:1}
\end{table*}  
\begin{table}[th!]
\centering
\begin{tabular}{@{} l *{3}{T{1.2}T{1.2}} T{1.2}T{1.2} @{}}
\toprule
& \multicolumn{2}{c}{CI tagging (Acc.)} \\
\cmidrule(lr){2-3} 
& {Baseline} & {CP retrieval}  \\
\midrule
\texttt{Llama2-13B} & 36.07 & \textbf{51.82} \\
\texttt{Llama2-70B} &  43.29 & \textbf{55.95} \\
\bottomrule
\end{tabular}
\caption{Results of using complexity score for retrieval (CP retrieval) on the CI parameters tagging task using \texttt{Llama2} models. Best values are in boldface.\vspace{-8mm}} 
\label{tab:2}
\end{table}
\begin{figure*}
    \centering
    \begin{subfigure}{0.32\textwidth}
        \includegraphics[width=\linewidth]{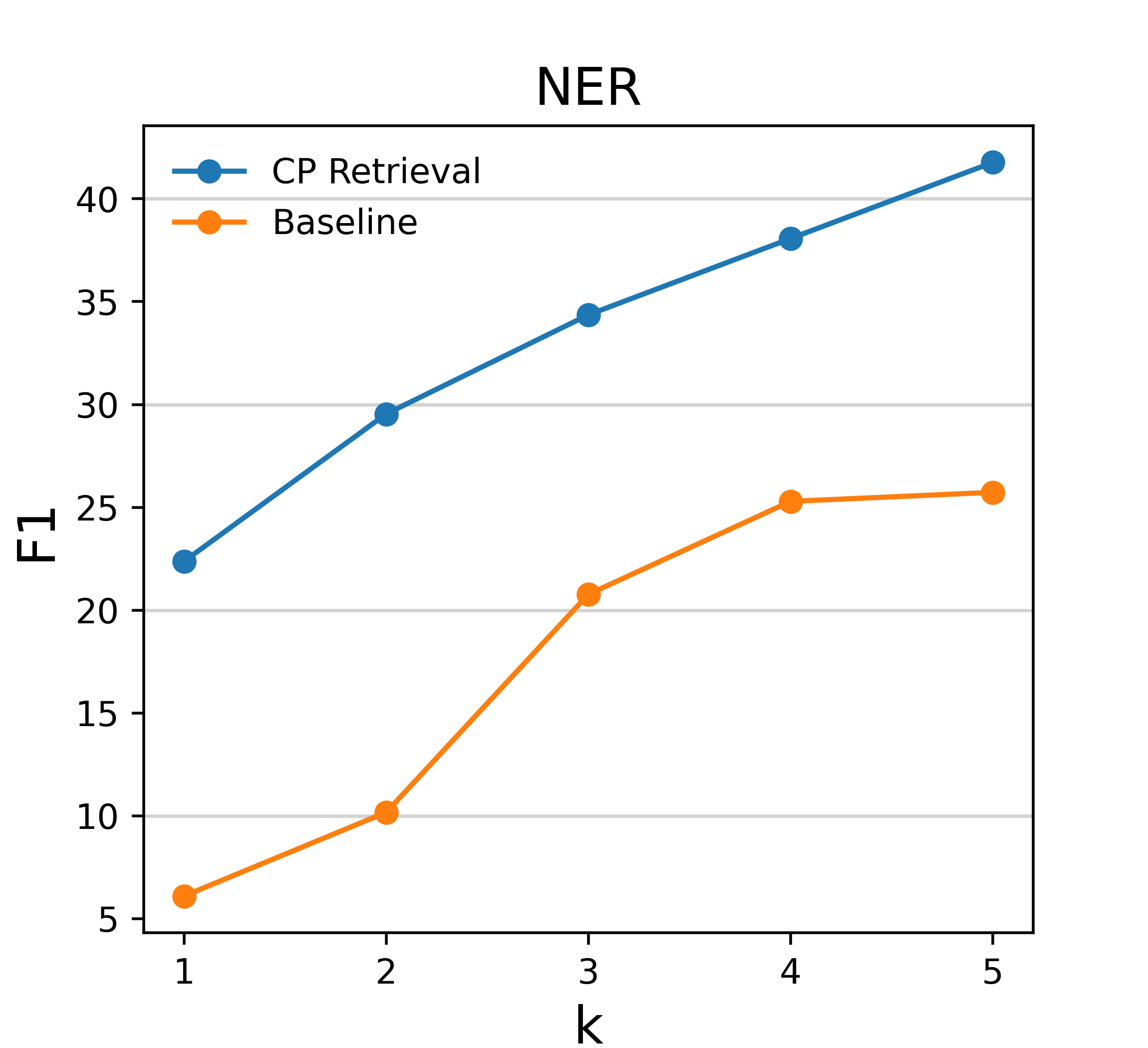}
        \caption{}
    \end{subfigure}
    \begin{subfigure}{0.32\textwidth}
        \includegraphics[width=\linewidth]{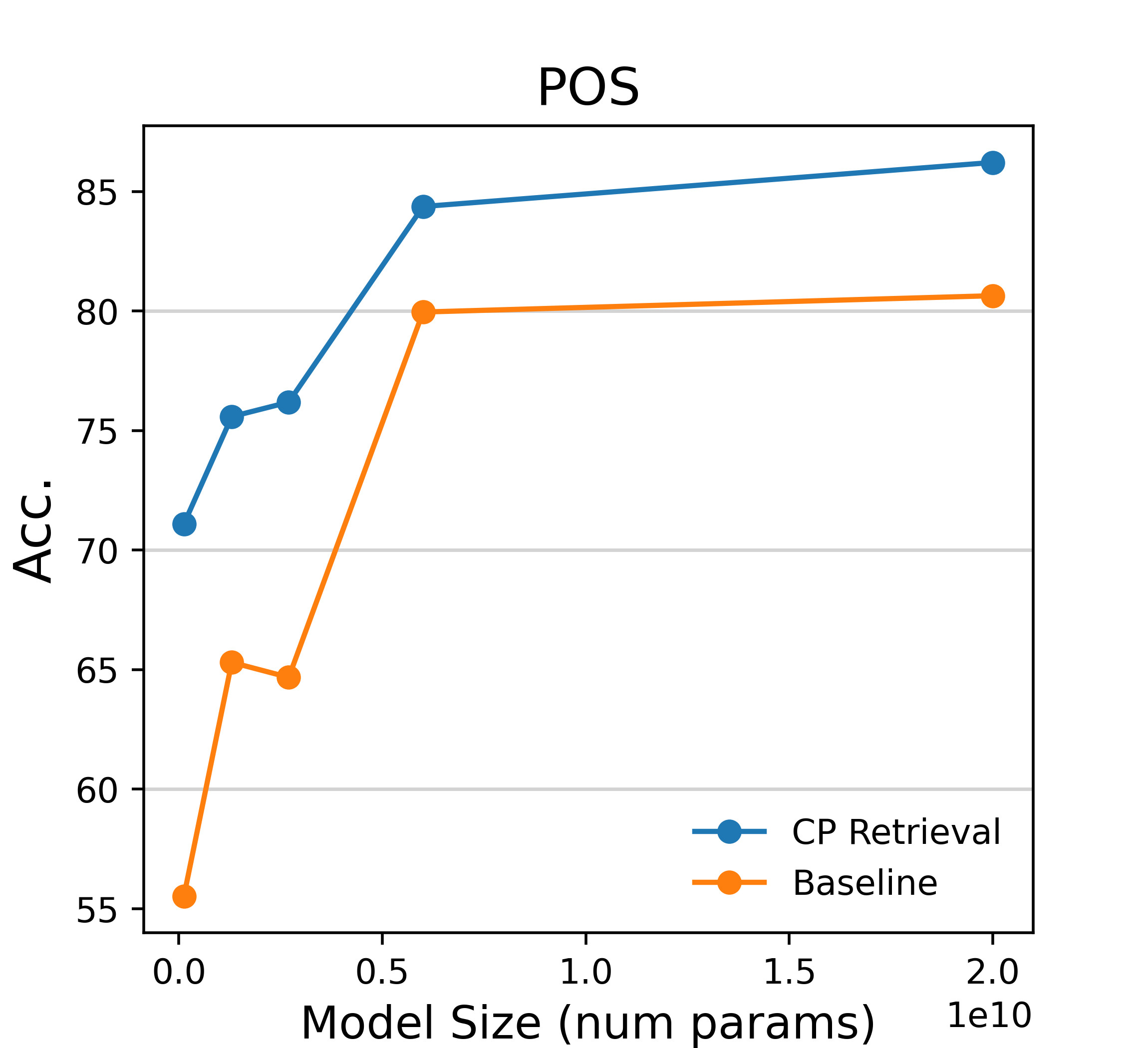}
        \caption{}
    \end{subfigure}
    \begin{subfigure}{0.32\textwidth}
        \includegraphics[width=\linewidth]{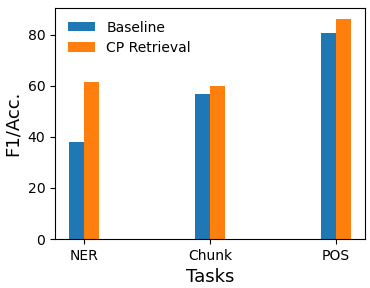}
        \caption{}
    \end{subfigure}
        \caption{{\bf CP retrieval demonstrates superior performance across all tasks.} (a) Performance while varying number of demonstrations $k$ (\texttt{GPT-Neo-2.7B}) for NER. (b) Performance while varying model size for a fixed $k$ = 5 for POS. (c) Performance for the various tasks on \texttt{GPT-NeoX-20B} model with $k$ = 5. Note that NER and chunking are evaluated based on the F1 score, and POS is evaluated based on accuracy as commonly done.\vspace{-2mm}}
    \label{fig:1}
\end{figure*}

\section{Results}
\label{sec:results}
Table~\ref{tab:1} compares the F1 and Accuracy scores (for NER, chunking and POS tagging) for different models while keeping the number of examples constant. Table~\ref{tab:2} provides the same for the CI tagging task on \texttt{Llama} models (performance in a different domain). Our results show that CP retrieval significantly improves few-shot accuracy over the baseline across all models and considered tasks. 

We observe substantial gains using our method, with the largest gains on \texttt{GPT-j-6B} which achieves a 28.85 point increase on NER, followed by \texttt{GPT-Neo-2.7B} which achieves a 24.82 point increase in chunking, and finally with \texttt{GPT-Neo-125M} which achieves a 15.56 point increase in POS tagging. It is also evident that the percentage gains of example selection drops with increasing model size in Fig.~\ref{fig:1}(b) (with \texttt{GPT-4} having the least gains), and this is due to CP retrieval approaching overall accuracy saturation. Low gains in \texttt{GPT-4} are also because larger models have seen more diverse training data, so random prompt selections are more than likely to cover the needed distribution.
However, even with small gains, we achieve state-of-the-art performance on the CoNLL2003 dataset with a 5\% increase over the previous state-of-the-art by~\citet{ashok2023promptner} for few-shot NER. 
\begin{table*}[th!]
\centering
\scriptsize
\resizebox{0.85\textwidth}{!}{
\begin{tabular}{@{} l *{3}{T{1.2}} @{}}
\toprule
& \multicolumn{1}{c}{NER (F1)} 
& \multicolumn{1}{c@{}}{Chunking (F1)} 
& \multicolumn{1}{c@{}}{POS (Acc.)}\\
\midrule
\texttt{Normalized Sentence Similarity} & 25.39 & 38.16 & 66.82 \\
\texttt{Normalized Smoothed Length Similarity}  &  17.22 & 31.50 & 59.91 \\
\texttt{Normalized Label Entropy}  & 15.08  & 28.11 & 60.43\\
\texttt{Baseline}  & 12.46 & 24.28 &  55.53\\
\bottomrule
\end{tabular}}
\caption{\textbf{Results of performing ablations to analyze the performance of each individual metric when used for retrieval of $k$ examples for a few-shot prompt (here $k$ = 5).}}
\label{tab:Ablation}
\end{table*}

\begin{figure*}
    \centering
    \begin{subfigure}{0.32\textwidth}
        \includegraphics[width=\linewidth]{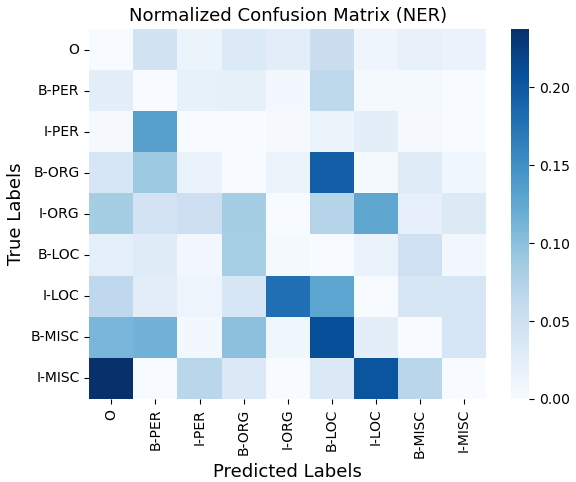}
        \caption{}
    \end{subfigure}
    \begin{subfigure}{0.33\textwidth}
        \includegraphics[width=\linewidth]{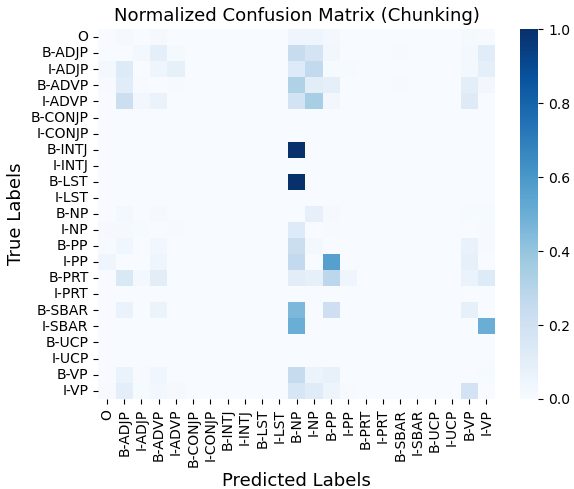}
        \caption{}
    \end{subfigure}
    \begin{subfigure}{0.33\textwidth}
        \includegraphics[width=\linewidth]{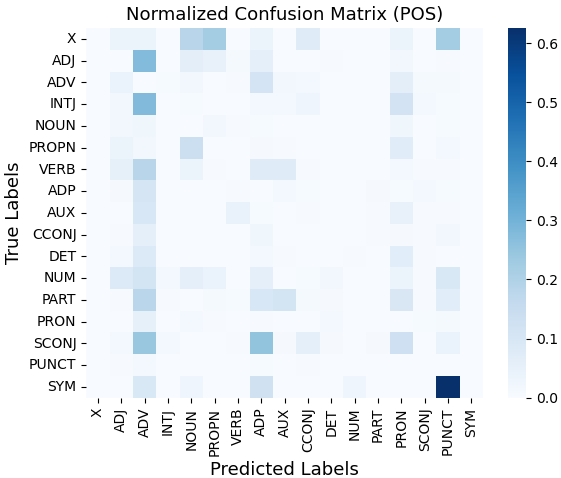}
        \caption{}
    \end{subfigure}
        \caption{\textbf{Normalized confusion matrices for (a) NER (b) Chunking and (c) POS}}
    \label{fig:ErrorAnalysis}
\end{figure*}
\\
\noindent{\bf Impact of $k$:} From Fig.~\ref{fig:1}(a), we see that our approach enables effective learning (larger gains) with smaller $k$. At $k$ = 2, we see a gain of 19.37 points over the baseline which decreased when $k$ is increased. The percentage gains consistently reduce as we increase the number of examples.

\noindent{\bf Discussion:}
What we want to put forth is that this approach can further boost performance when combined with more sophisticated prompting techniques. For example, CP retrieval can be used along with chain-of-thought (CoT) prompting~\citep{wei2023chainofthought}. It can be used to identify the most information-rich examples for the initial context in CoT prompting.
\subsection{Ablation Study}
\noindent{\bf Impact of Indvidual Complexity Metrics:} Table~\ref{tab:Ablation} shows the performance of each individual metric in isolation to retrieve the best $k$ examples for the few-shot prompt. It is evident that all 3 metrics lead to gains in accuracy/F1 scores for each of these datasets. It is also important to observe that the Normalized Sentence Similarity metric leads to the highest gains in performance over the baseline. This is also directly reflected in the optimal values for the weights $(w_1,w_2,w_3)$ in seen in \S~\ref{sec:ComplexityScore} which have a higher value associated with $w_3$. 

\noindent{\bf Other Baselines:} We also compared our methodology with the $k$NN baseline in Appendix~\ref{app:baseline} demonstrating that CP retrieval consistently outperforms $k$NN-based example retrieval. For example, it results in a 4.27\% absolute gain in F1 score over the $k$NN baseline for the task of sentence chunking when using \texttt{GPT-Neo-2.7B} model.

\noindent{\bf Number of Samples:} In Appendix~\ref{app:num_samples}, we show that CP retrieval helps in improving performance {\em even when} the number of samples to retrieve from in the train set is reduced. Specifically, when the number of available samples for the NER task is reduced to 1000 (from 11079), CP retrieval achieves an F1 score of 25.98 compared to the baseline random selection score of 12.46. Even with just 30 samples, CP retrieval attains an F1 score of 18.19.

\subsection{Error Analysis}

In this subsection, we analyze the common errors made by PLMs using our methodology to identify patterns and potential areas for improvement. We do so by looking at the confusion matrices shown in Fig.~\ref{fig:ErrorAnalysis}. To plot these, we have zeroed out the diagonal values which represent correct predictions and then perform normalization, making it easier to compare error rates across different classes.\\
We see that error rate is more evenly spread out for NER (with a max of around 0.25)
(c.f. Fig.~\ref{fig:ErrorAnalysis}(a)), whereas Chunking (c.f. Fig.~\ref{fig:ErrorAnalysis}(b)) and POS (c.f. Fig.~\ref{fig:ErrorAnalysis}(c)) have more localized errors. We see vertical bands in the confusion matrices of Chunking and POS, and these vertical bands can indicate systematic biases or recurring mistakes in the model which can be targeted for correction. Specifically, the model over predicts many words as "B-NP" (Beginning of a Noun Phrase) in chunking and incorrectly predicts the "SYM" (Symbol) tag as "PUNCT" (Punctuation) in POS tagging.  
\section{Conclusions}

In summary, CP retrieval is a flexible technique that can be used to enhance the accuracy of existing few-shot approaches using example scoring. Additionally, it provides a task-agnostic quantification of the informativeness of examples for improved model performance. The relative weights of the components of the complexity score can be tuned on a per-task basis. Our approach allows us to extract the most out PLMs without any fine-tuning of model parameters.
\newpage
\section{Limitations}

Our methodology focuses on sequence tagging tasks and cannot be applied to tasks outside this domain such as question answering. This is because the normalized entropy metric is specifically designed for encouraging selection of sentences with a wider label variety on the token level; this helps increase the sampling of underrepresented labels. Further, we have conducted all experiments for the sequence tagging tasks in only the English language. Performance with other languages has not been analyzed.

\bibliography{custom}

\begin{thebibliography}{17}
\expandafter\ifx\csname natexlab\endcsname\relax\def\natexlab#1{#1}\fi

\bibitem[{Ashok and Lipton(2023)}]{ashok2023promptner}
Dhananjay Ashok and Zachary~C. Lipton. 2023.
\newblock \href {http://arxiv.org/abs/2305.15444} {Promptner: Prompting for named entity recognition}.

\bibitem[{Blevins et~al.(2023)Blevins, Gonen, and Zettlemoyer}]{blevins2023prompting}
Terra Blevins, Hila Gonen, and Luke Zettlemoyer. 2023.
\newblock \href {http://arxiv.org/abs/2211.07830} {Prompting language models for linguistic structure}.

\bibitem[{Brown et~al.(2020)Brown, Mann, Ryder, Subbiah, Kaplan, Dhariwal, Neelakantan, Shyam, Sastry, Askell, Agarwal, Herbert-Voss, Krueger, Henighan, Child, Ramesh, Ziegler, Wu, Winter, Hesse, Chen, Sigler, Litwin, Gray, Chess, Clark, Berner, McCandlish, Radford, Sutskever, and Amodei}]{brown2020language}
Tom~B. Brown, Benjamin Mann, Nick Ryder, Melanie Subbiah, Jared Kaplan, Prafulla Dhariwal, Arvind Neelakantan, Pranav Shyam, Girish Sastry, Amanda Askell, Sandhini Agarwal, Ariel Herbert-Voss, Gretchen Krueger, Tom Henighan, Rewon Child, Aditya Ramesh, Daniel~M. Ziegler, Jeffrey Wu, Clemens Winter, Christopher Hesse, Mark Chen, Eric Sigler, Mateusz Litwin, Scott Gray, Benjamin Chess, Jack Clark, Christopher Berner, Sam McCandlish, Alec Radford, Ilya Sutskever, and Dario Amodei. 2020.
\newblock \href {http://arxiv.org/abs/2005.14165} {Language models are few-shot learners}.

\bibitem[{Chen et~al.(2023)Chen, Zaharia, and Zou}]{chen2023chatgpts}
Lingjiao Chen, Matei Zaharia, and James Zou. 2023.
\newblock \href {http://arxiv.org/abs/2307.09009} {How is chatgpt's behavior changing over time?}

\bibitem[{Liu et~al.(2022)Liu, Shen, Zhang, Dolan, Carin, and Chen}]{liu-etal-2022-makes}
Jiachang Liu, Dinghan Shen, Yizhe Zhang, Bill Dolan, Lawrence Carin, and Weizhu Chen. 2022.
\newblock \href {https://doi.org/10.18653/v1/2022.deelio-1.10} {What makes good in-context examples for {GPT}-3?}
\newblock In \emph{Proceedings of Deep Learning Inside Out (DeeLIO 2022): The 3rd Workshop on Knowledge Extraction and Integration for Deep Learning Architectures}, pages 100--114, Dublin, Ireland and Online. Association for Computational Linguistics.

\bibitem[{McHugh(2012)}]{mchugh2012interrater}
Mary~L McHugh. 2012.
\newblock Interrater reliability: the kappa statistic.
\newblock \emph{Biochemia medica}, 22(3):276--282.

\bibitem[{Min et~al.(2022)Min, Lyu, Holtzman, Artetxe, Lewis, Hajishirzi, and Zettlemoyer}]{min2022rethinking}
Sewon Min, Xinxi Lyu, Ari Holtzman, Mikel Artetxe, Mike Lewis, Hannaneh Hajishirzi, and Luke Zettlemoyer. 2022.
\newblock \href {http://arxiv.org/abs/2202.12837} {Rethinking the role of demonstrations: What makes in-context learning work?}

\bibitem[{Nivre et~al.(2020)Nivre, de~Marneffe, Ginter, Haji{\v{c}}, Manning, Pyysalo, Schuster, Tyers, and Zeman}]{nivre-etal-2020-universal}
Joakim Nivre, Marie-Catherine de~Marneffe, Filip Ginter, Jan Haji{\v{c}}, Christopher~D. Manning, Sampo Pyysalo, Sebastian Schuster, Francis Tyers, and Daniel Zeman. 2020.
\newblock \href {https://aclanthology.org/2020.lrec-1.497} {{U}niversal {D}ependencies v2: An evergrowing multilingual treebank collection}.
\newblock In \emph{Proceedings of the Twelfth Language Resources and Evaluation Conference}, pages 4034--4043, Marseille, France. European Language Resources Association.

\bibitem[{Rubin et~al.(2022)Rubin, Herzig, and Berant}]{rubin2022learning}
Ohad Rubin, Jonathan Herzig, and Jonathan Berant. 2022.
\newblock \href {http://arxiv.org/abs/2112.08633} {Learning to retrieve prompts for in-context learning}.

\bibitem[{Sclar et~al.(2024)Sclar, Choi, Tsvetkov, and Suhr}]{sclar2024quantifying}
Melanie Sclar, Yejin Choi, Yulia Tsvetkov, and Alane Suhr. 2024.
\newblock \href {https://openreview.net/forum?id=RIu5lyNXjT} {Quantifying language models' sensitivity to spurious features in prompt design or: How i learned to start worrying about prompt formatting}.
\newblock In \emph{The Twelfth International Conference on Learning Representations}.

\bibitem[{Shvartzshnaider et~al.(2019)Shvartzshnaider, Apthorpe, Feamster, and Nissenbaum}]{Shvartzshnaider2019GoingAT}
Yan Shvartzshnaider, Noah~J. Apthorpe, Nick Feamster, and Helen Nissenbaum. 2019.
\newblock \href {https://api.semanticscholar.org/CorpusID:210874267} {Going against the (appropriate) flow: A contextual integrity approach to privacy policy analysis}.
\newblock In \emph{AAAI Conference on Human Computation \& Crowdsourcing}.

\bibitem[{Sorensen et~al.(2022)Sorensen, Robinson, Rytting, Shaw, Rogers, Delorey, Khalil, Fulda, and Wingate}]{Sorensen_2022}
Taylor Sorensen, Joshua Robinson, Christopher Rytting, Alexander Shaw, Kyle Rogers, Alexia Delorey, Mahmoud Khalil, Nancy Fulda, and David Wingate. 2022.
\newblock \href {https://doi.org/10.18653/v1/2022.acl-long.60} {An information-theoretic approach to prompt engineering without ground truth labels}.
\newblock In \emph{Proceedings of the 60th Annual Meeting of the Association for Computational Linguistics (Volume 1: Long Papers)}. Association for Computational Linguistics.

\bibitem[{Tjong Kim~Sang and Buchholz(2000)}]{tjong-kim-sang-buchholz-2000-introduction}
Erik~F. Tjong Kim~Sang and Sabine Buchholz. 2000.
\newblock \href {https://aclanthology.org/W00-0726} {Introduction to the {C}o{NLL}-2000 shared task chunking}.
\newblock In \emph{Fourth Conference on Computational Natural Language Learning and the Second Learning Language in Logic Workshop}.

\bibitem[{Tjong Kim~Sang and De~Meulder(2003)}]{tjong-kim-sang-de-meulder-2003-introduction}
Erik~F. Tjong Kim~Sang and Fien De~Meulder. 2003.
\newblock \href {https://aclanthology.org/W03-0419} {Introduction to the {C}o{NLL}-2003 shared task: Language-independent named entity recognition}.
\newblock In \emph{Proceedings of the Seventh Conference on Natural Language Learning at {HLT}-{NAACL} 2003}, pages 142--147.

\bibitem[{Wei et~al.(2023)Wei, Wang, Schuurmans, Bosma, Ichter, Xia, Chi, Le, and Zhou}]{wei2023chainofthought}
Jason Wei, Xuezhi Wang, Dale Schuurmans, Maarten Bosma, Brian Ichter, Fei Xia, Ed~Chi, Quoc Le, and Denny Zhou. 2023.
\newblock \href {http://arxiv.org/abs/2201.11903} {Chain-of-thought prompting elicits reasoning in large language models}.

\bibitem[{Wilson et~al.(2016)Wilson, Schaub, Dara, Liu, Cherivirala, Giovanni~Leon, Schaarup~Andersen, Zimmeck, Sathyendra, Russell, Norton, Hovy, Reidenberg, and Sadeh}]{wilson-etal-2016-creation}
Shomir Wilson, Florian Schaub, Aswarth~Abhilash Dara, Frederick Liu, Sushain Cherivirala, Pedro Giovanni~Leon, Mads Schaarup~Andersen, Sebastian Zimmeck, Kanthashree~Mysore Sathyendra, N.~Cameron Russell, Thomas~B. Norton, Eduard Hovy, Joel Reidenberg, and Norman Sadeh. 2016.
\newblock \href {https://doi.org/10.18653/v1/P16-1126} {The creation and analysis of a website privacy policy corpus}.
\newblock In \emph{Proceedings of the 54th Annual Meeting of the Association for Computational Linguistics (Volume 1: Long Papers)}, pages 1330--1340, Berlin, Germany. Association for Computational Linguistics.

\bibitem[{Zeldes(2017)}]{10.1007/s10579-016-9343-x}
Amir Zeldes. 2017.
\newblock \href {https://doi.org/10.1007/s10579-016-9343-x} {The gum corpus: creating multilayer resources in the classroom}.
\newblock \emph{Lang. Resour. Eval.}, 51(3):581–612.

\end{thebibliography}

\onecolumn
\newpage

\appendix

\section*{Appendix}
\label{sec:appendix}
\section{Prompts}
\label{app:prompts}
\begin{figure*}[!th]
\begin{tcolorbox}[
    top=1mm, 
    bottom=1mm, 
    left=1mm, 
    right=1mm,
    sharp corners,
    ]    
\begin{small}
\textbf{Context}: 
The Finance Ministry raised the price for tap sales of the Dutch government 's new 5.75 percent bond due September 2002 to 99.95 from 99.90 .
\par
\textbf{Tagged}: The\_O Finance\_B-ORG Ministry\_I-ORG raised\_O the\_O price\_O for\_O tap\_O sales\_O of\_O the\_O Dutch\_B-MISC government\_O 's\_O new\_O 5.75\_O percent\_O bond\_O due\_O September\_O 2002\_O to\_O 99.95\_O from\_O 99.90\_O .\_O
\par
\textbf{Context}: Swiss bonds ended mostly higher in generally quiet activity , with the September confederate bond futures contract holding just above 113.00 .
\par
\textbf{Tagged}: Swiss\_B-MISC bonds\_O ended\_O mostly\_O higher\_O in\_O generally\_O quiet\_O activity\_O ,\_O with\_O the\_O September\_O confederate\_O bond\_O futures\_O contract\_O holding\_O just\_O above\_O 113.00\_O .\_O
\par
\textbf{Context}: The Brent crude futures market on the Singapore International Monetary Exchange ( SIMEX ) was closed on Monday in respect for a U.K. national holiday .
\par
\textbf{Tagged}: The\_O Brent\_B-ORG crude\_O futures\_O market\_O on\_O the\_O Singapore\_B-ORG International\_I-ORG Monetary\_I-ORG Exchange\_I-ORG (\_O SIMEX\_B-ORG )\_O was\_O closed\_O on\_O Monday\_O in\_O respect\_O for\_O a\_O U.K.\_B-LOC national\_O holiday\_O .\_O\par
\textbf{Context}: European bourses closed mixed on Tuesday with London clawing back most of the day 's losses despite an unsteady start on wall Street , hit by inflation worries .
\par
\textbf{Tagged}: European\_B-MISC bourses\_O closed\_O mixed\_O on\_O Tuesday\_O with\_O London\_B-LOC clawing\_O back\_O most\_O of\_O the\_O day\_O 's\_O losses\_O despite\_O an\_O unsteady\_O start\_O on\_O wall\_B-ORG Street\_I-ORG ,\_O hit\_O by\_O inflation\_O worries\_O .\_O
\par
\textbf{Context}: No closures of airports in the Commonwealth of Independent States are expected on August 24 and August 25 , the Russian Weather Service said on Friday .
\par
\textbf{Tagged}: No\_O closures\_O of\_O airports\_O in\_O the\_O Commonwealth\_B-LOC of\_I-LOC Independent\_I-LOC States\_I-LOC are\_O expected\_O on\_O August\_O 24\_O and\_O August\_O 25\_O ,\_O the\_O Russian\_B-ORG Weather\_I-ORG Service\_I-ORG said\_O on\_O Friday\_O .\_O
\par
\textbf{Context: Hungarian overnight interest rates closed higher on Friday as market liquidity tightened before the December 10 social security contribution payment deadline, dealers said.}\par
\textbf{Tagged}:
\par
\end{small}
\end{tcolorbox}
\caption{Prompt generated for a specific test sample after retrieval of 5 examples (5-shot prompt) from the training dataset for \textbf{Named Entity Recognition} using our approach. The test sentence that we want the model to label is provided at the end of the prompt and is shown in bold}
\label{tab:prompt1}
\end{figure*}
\begin{figure*}[!th]
\begin{tcolorbox}[
    top=1mm, 
    bottom=1mm, 
    left=1mm, 
    right=1mm,
    sharp corners,
    ]    
\begin{small}
\textbf{Context:} Shearson is offering the notes as 6 3\/4 \% securities priced to yield 6.15 \% .\par
\textbf{Tagged:} Shearson\_B-NP is\_B-VP offering\_I-VP the\_B-NP notes\_I-NP as\_B-PP 6\_B-NP 3\/4\_I-NP \%\_I-NP securities\_I-NP priced\_B-VP to\_B-VP yield\_I-VP 6.15\_B-NP \%\_I-NP .\_O\par
\textbf{Context:} Bonds due 1991-1996 carry 6.70 \% coupons and bonds due 1997-2000 carry 6 3\/4 \% coupons .\par
\textbf{Tagged:} Bonds\_B-NP due\_B-ADJP 1991-1996\_B-NP carry\_B-VP 6.70\_B-NP \%\_I-NP coupons\_I-NP and\_O bonds\_B-NP due\_B-ADJP 1997-2000\_B-NP carry\_B-VP 6\_B-NP 3\/4\_I-NP \%\_I-NP coupons\_I-NP .\_O\par
\textbf{Context:} Their price falls further than that of other bonds when inflation and interest rates kick up .\par
\textbf{Tagged:} Their\_B-NP price\_I-NP falls\_B-VP further\_B-ADVP than\_B-PP that\_B-NP of\_B-PP other\_B-NP bonds\_I-NP when\_B-ADVP inflation\_B-NP and\_O interest\_B-NP rates\_I-NP kick\_B-VP up\_B-PRT .\_O\par
\textbf{Context:} Serial bonds were priced at par to yield to 6.90 \% in 2000 .\par
\textbf{Tagged:} Serial\_B-NP bonds\_I-NP were\_B-VP priced\_I-VP at\_B-PP par\_B-NP to\_B-VP yield\_I-VP to\_B-PP 6.90\_B-NP \%\_I-NP in\_B-PP 2000\_B-NP .\_O\par
\textbf{Context:} At the auction of six-month U.S. Treasury bills on Monday , the average yield fell to 7.61 \% from 7.82 \% .\par
\textbf{Tagged:} At\_B-PP the\_B-NP auction\_I-NP of\_B-PP six-month\_B-NP U.S.\_I-NP Treasury\_I-NP bills\_I-NP on\_B-PP Monday\_B-NP ,\_O the\_B-NP average\_I-NP yield\_I-NP fell\_B-VP to\_B-PP 7.61\_B-NP \%\_I-NP from\_B-PP 7.82\_B-NP \%\_I-NP .\_O\par
\textbf{Context: The rate on six-month bills rose to 7.53 \% for a bond-equivalent yield of 7.92 \% .}\par
\textbf{Tagged:}\par
\end{small}
\end{tcolorbox}
\caption{Prompt generated for a specific test sample after retrieval of 5 examples (5-shot prompt) from the training dataset for \textbf{Sentence Chunking} using our approach. The test sentence that we want the model to label is provided at the end of the prompt and is shown in bold}
\label{tab:prompt2}
\end{figure*}
\begin{figure*}[!th]
\begin{tcolorbox}[
    top=1mm, 
    bottom=1mm, 
    left=1mm, 
    right=1mm,
    sharp corners,
    ]    
\begin{small}
\textbf{Context:} Many forms of culture are passed down through a combination of deliberate and unconscious processes .\par
\textbf{Tagged:} Many\_ADJ forms\_NOUN of\_ADP culture\_NOUN are\_AUX passed\_VERB down\_ADP through\_ADP a\_DET combination\_NOUN of\_ADP deliberate\_ADJ and\_CCONJ unconscious\_ADJ processes\_NOUN .\_PUNCT\par
\textbf{Context:} Cognitive psychology is the field of psychology dedicated to examining how people think .\par
\textbf{Tagged:} Cognitive\_ADJ psychology\_NOUN is\_AUX the\_DET field\_NOUN of\_ADP psychology\_NOUN dedicated\_VERB to\_SCONJ examining\_VERB how\_ADV people\_NOUN think\_VERB .\_PUNCT\par
\textbf{Context:} Having been a psychologist for a number of years gives me a leg up on it .\par
\textbf{Tagged:} Having\_AUX been\_AUX a\_DET psychologist\_NOUN for\_ADP a\_DET number\_NOUN of\_ADP years\_NOUN gives\_VERB me\_PRON a\_DET leg\_NOUN up\_ADV on\_ADP it\_PRON .\_PUNCT\par
\textbf{Context:} Do they get mad or irritated if the centre of attention moves to someone else ?\par
\textbf{Tagged:} Do\_AUX they\_PRON get\_VERB mad\_ADJ or\_CCONJ irritated\_VERB if\_SCONJ the\_DET centre\_NOUN of\_ADP attention\_NOUN moves\_VERB to\_ADP someone\_PRON else\_ADV ?\_PUNCT\par
\textbf{Context:} When thoughts are formed , the brain also pulls information from emotions and memories ( Figure 7.2 ) .\par
\textbf{Tagged:} When\_ADV thoughts\_NOUN are\_AUX formed\_VERB ,\_PUNCT the\_DET brain\_NOUN also\_ADV pulls\_VERB information\_NOUN from\_ADP emotions\_NOUN and\_CCONJ memories\_NOUN (\_PUNCT Figure\_PROPN 7.2\_NUM )\_PUNCT .\_PUNCT\par
\textbf{Context: Psychologists have examined the mental processes that underpin conscious and unconscious biases [ 6 ] ;}\par
\textbf{Tagged}:\par
\end{small}
\end{tcolorbox}
\caption{Prompt generated for a specific test sample after retrieval of 5 examples (5-shot prompt) from the training dataset for \textbf{Part-of-Speech} tagging using our approach. The test sentence that we want the model to label is provided at the end of the prompt and is shown in bold}
\label{tab:prompt3}
\end{figure*}
\begin{figure*}[!th]
\begin{tcolorbox}[
    top=1mm, 
    bottom=1mm, 
    left=1mm, 
    right=1mm,
    sharp corners,
    ]    
\begin{small}
\textbf{Context}: Classify each word in the the sentence to its contextual integrity parameter (Sender, Attribute, Receiver, None, TP, Subject) \par
\textbf{Example1}:\par
\textbf{Input Sentence}:\par
[We, will, also, disclose, nonpersonally, identifiable, information, to, our, partners, and, other, third, parties, about, how, our, users, collectively, use, the, Sites]\par
\textbf{Output}:\par
[[We, Sender], [will, None], [also, None], [disclose, None], [nonpersonally, Attribute], [identifiable, Attribute], [information, Attribute], [to, None], [our, Receiver], [partners, Receiver], [and, None], [other, Receiver], [third, Receiver], [parties, Receiver], [about, TP], [how, TP], [our, TP], [users, TP], [collectively, TP], [use, TP], [the, TP], [Sites, TP]] \par        
\textbf{Example2}:\par
\textbf{Input Sentence}:\par
[If, you, choose, to, use, our, referral, service, to, tell, a, friend, about, Military, or, refer, other, information, on, Military, to, a, friend, we, will, ask, you, for, your, friends, name, and, email, address] \par
\textbf{Output}:\par
[[If, TP], [you, TP], [choose, TP], [to, TP], [use, TP], [our, TP], [referral, TP], [service, TP], [to, TP], [tell, TP], [a, TP], [friend, TP], [about, TP], [Military, Attribute], [or, TP], [refer, TP], [other, Attribute], [information, Attribute], [on, TP], [Military, Attribute], [to, TP], [a, TP], [friend, TP], [we, Receiver], [will, None], [ask, None], [you, Sender], [for, None], [your, Subject], [friends, Subject], [name, Attribute], [and, None], [email, Attribute], [address, Attribute]] \par
\textbf{Question}: Now perform the same task on the input given below and provide output in the same format as above:\par
\textbf{Input Sentence}: \par
[Insert Input] \par
\end{small}
\end{tcolorbox}
\caption{Prompt structure for the \textbf{CI} tagging task (2-shot prompt shown here)}
\label{tab:prompt3}
\end{figure*}
\newpage
\section{\texorpdfstring{$k$}{k}NN Baseline}
\label{app:baseline}
\begin{table*}[th!]
\centering
\resizebox{\textwidth}{!}{
\begin{tabular}{@{} l *{3}{T{1.2}T{1.2}} T{1.2}T{1.2} @{}}
\toprule
& \multicolumn{2}{c}{NER (F1)} 
& \multicolumn{2}{c@{}}{Chunking (F1)} 
& \multicolumn{2}{c@{}}{POS (Acc.)}\\
\cmidrule(lr){2-3} \cmidrule(l){4-5} \cmidrule(l){6-7}
& {$k$NN Baseline} & {CP retrieval} & {$k$NN Baseline} & {CP retrieval} & {$k$NN Baseline} & {CP retrieval}\\
\midrule
\texttt{GPT-Neo-125M} & 25.39 & \textbf{29.21} & 38.16 & \textbf{40.33} & 66.82 & \textbf{71.09}\\
\texttt{GPT-Neo-2.7B}  & 36.44 & \textbf{41.77} & 49.05 & \textbf{53.32} & 73.71 & \textbf{76.19}\\
\bottomrule
\end{tabular}}
\caption{ Results of using complexity score retrieval (CP retrieval) of the best scoring $k$ examples for the few-shot prompt (here $k$ = 5). The baseline here is picking the $k$-nearest neighbors + structured prompting paradigm \citep{blevins2023prompting}.  Best values are in boldface which demonstrates that CP retrieval aids in significant performance gains.}
\label{tab:3}
\end{table*}
\section{Contextual Integrity Dataset Details}
\label{app:CI}

The Contextual Integrity (CI) parameter tagging task~\citep{Shvartzshnaider2019GoingAT} involves classifying the words in relevant sentences of privacy policies into the following classes - Sender, Receiver, Subject, Attribute and Transmission Principle. The total counts for each of these classes in the dataset is shown in Table~\ref{tab:table_classcounts}. The dataset has a total of 600 samples (sentences) which have been extracted from The OPP-115 Corpus (Online Privacy Policies, set of 115)~\citep{wilson-etal-2016-creation} and labelled for this sequence tagging task. We utilize \texttt{GPT-4} to create this dataset in a few-shot scenario (the examples for the few-shot prompts are manually labelled). Following this, we obtain the labelled outputs from \texttt{GPT-4} which is then passed through human verification and relabelling. 3 annotators performed this task such that each sample is verified by 2 annotators and we achieve an average inter-annotator agreement of 0.64 using the popular Cohen’s Kappa metric~\citep{mchugh2012interrater} on the relabelled classes.
We have used 100 samples in the test set and the remaining 500 samples as part of the train set. 
\begin{table}[h]
    \centering
    \begin{tabular}{lc}
        \hline
        \textbf{Class} & \textbf{Frequency} \\
        \hline
        O & 4817\\
        SENDER & 481\\
        RECEIVER & 1058\\
        SUBJECT & 509\\
        ATTRIBUTE & 3152\\
        TP & 6798\\
        \hline
    \end{tabular}
    \vspace{10pt}
    \caption{Total class counts in the CI dataset}
    \label{tab:table_classcounts}
\end{table}
\section{Dependence on Number of Training Samples Available for Retrieval}
\label{app:num_samples}

While most practical tasks nowadays usually have a large set of examples present to select from, we wanted to check the viability of our methodology  as the number of examples for selection reduced. The experiment was performed on \texttt{GPT-Neo-125M} with a 5-shot prompt on the NER datatset. It is evident from Table~\ref{tab:table_trainingsetsize} that reducing the number of examples for selection reduces the F1 score gradually, but there are still significant gains even with a small set of examples to select from when it is compared to the random selection baseline.
\begin{table}[h]
    \centering
    \begin{tabular}{lc}
        \hline
        \textbf{No. of samples to select from} & \textbf{CP retrieval (F1)} \\
        \hline
        11079 (Entire train data)  & 29.21\\
        1000  & 25.98\\
        500 & 24.38\\
        100  & 23.74\\
        30  & 18.19\\
        \hline
         & \textbf{Baseline-random sample selection (F1)}\\
        \hline
        \vspace{1pt}
        11079 (Entire train data) & 12.46\\
        \hline
    \end{tabular}
    \vspace{10pt}
    \caption{Performance when number of samples to select from (train set) is reduced gradually. Even with just 30 samples to select from, CP retrieval provides a 5.73\% absolute gain in F1 score.}
    \label{tab:table_trainingsetsize}
\end{table}
\end{document}